%% file: main.tex
\documentclass[sigconf]{acmart}

\AtBeginDocument{%
  }

\setcopyright{none}
\copyrightyear{2026}
\acmYear{2026}
\acmDOI{XXXXXXX.XXXXXXX}
\acmConference[Agent4IR @ KDD '26]{KDD 2026 Workshop on AI Agent for Information Retrieval}{August 9--13, 2026}{Jeju, Republic of Korea}
\acmISBN{978-1-4503-XXXX-X/2026/08}

\usepackage{color, xspace}
\usepackage{enumitem}

\title{Beyond Benchmark Islands: Toward Representative Trustworthiness Evaluation for Agentic AI}

\author{Jinhu Qi}
\email{jhqi25@cse.cuhk.edu.hk}
\affiliation{%
  \institution{The Chinese University of Hong Kong}
  \country{Hong Kong SAR, China}}

\author{Yifan Li}
\email{liyifan@link.cuhk.edu.hk}
\affiliation{%
  \institution{The Chinese University of Hong Kong}
  \country{Hong Kong SAR, China}}

\author{Minghao Zhao}
\email{mhzhao@link.cuhk.edu.hk}
\affiliation{%
  \institution{The Chinese University of Hong Kong}
  \country{Hong Kong SAR, China}}

\author{Wentao Zhang}
\email{p2522808@mpu.edu.mo}
\affiliation{%
  \institution{Macao Polytechnic University}
  \country{Macao SAR, China}}

\author{Zijian Zhang}
\email{zhangzijian@jlu.edu.cn}
\affiliation{%
  \institution{Jilin University}
  \country{Changchun, China}}
  
\author{Yaoman Li}
\email{ymli@link.cuhk.edu.hk}
\affiliation{%
  \institution{The Chinese University of Hong Kong}
  \country{Hong Kong SAR, China}}

\author{Irwin King}
\email{king@cse.cuhk.edu.hk}
\affiliation{%
  \institution{The Chinese University of Hong Kong}
  \country{Hong Kong SAR, China}}

\renewcommand{\shortauthors}{Qi and Li, et al.}

\begin{abstract}
\input{Chapter/0_abstract}
\end{abstract}



\begin{document}

\maketitle

%% --- Chapter inputs ---
\input{Chapter/1_introduction}

\input{Chapter/2_related_work}

\input{Chapter/3_framework}

\input{Chapter/4_experiments}

\input{Chapter/5_discussion}
\input{Chapter/6_conclusion}

%%
%% Acknowledgments
%%
% \begin{acks}
% % TODO: Add acknowledgments here.
% \end{acks}

%%
%% Bibliography
%%
\bibliographystyle{ACM-Reference-Format}
\bibliography{reference}

\end{document}

%% file: Chapter/0_abstract.tex
Agentic AI systems are moving from static question answering to tool-augmented, multi-step workflows whose failures---unsafe tool use, unauthorised actions, social harm---carry deployment-level consequences. Yet evaluation practice remains fragmented across isolated benchmark slices, and the term ``trustworthiness'' is frequently invoked but rarely defined operationally.

We argue the central limitation is twofold: (i)~the absence of a concrete, measurable specification of \emph{what} agent trustworthiness means, and (ii)~the lack of a principled notion of \emph{representativeness} that would allow agent behaviour to be assessed over a socio-technical scenario distribution rather than a collection of disconnected benchmark instances. We address (i) by defining agentic trustworthiness as a five-property profile---Reliability, Robustness, Safety, Social-Ethical Alignment, and Operational Integrity---grounded in current AI risk frameworks; and we address (ii) with the \textbf{Holographic Agent Assessment Framework (HAAF)}, which characterises this profile over a \emph{scenario manifold} via four complementary components (static policy analysis, interactive sandbox simulation, social-ethical alignment assessment, and distribution-aware representative sampling) connected through an iterative \emph{Trustworthy Optimization Factory} that converts red-team diagnoses into blue-team interventions.

Our contributions are: \textbf{(1)} an operational, measurable definition of agentic trustworthiness through five properties; \textbf{(2)} a distribution-aware scenario-sampling framework that surfaces property-level trade-offs invisible to scalar leaderboards; and \textbf{(3)} a cross-family transfer experiment in which interventions designed from a single focal model's failures generalise---without per-model or per-scenario tuning---to 13 contemporary systems spanning seven model families (Llama, Mistral, Kimi, GLM, Qwen, GPT, DeepSeek) on a 100-scenario suite, where all 13 systems improve and two reach a perfect risk-weighted profile, establishing HAAF's Factory as a model-agnostic deployment-readiness pipeline. Code and data are available at \url{https://github.com/TonyQJH/haaf-pilot}.

%% file: Chapter/1_introduction.tex
\section{Introduction}
\label{sec:introduction}

Agentic AI systems are rapidly evolving from passive question-answering assistants into autonomous decision-makers that plan, call tools, and act in open-ended environments~\cite{mialon2023gaia,drouin2024workarena,xie2024osworld}. This shift is especially visible in information-retrieval (IR) contexts, where modern agents now act on retrieved documents and structured data---issuing follow-up queries, writing files, sending messages, and modifying records---rather than merely returning passages~\cite{xi2024agentir}. As these systems become embedded in real-world workflows, their failures are no longer limited to incorrect answers on static tasks. Instead, they may exhibit goal drift, unsafe tool use, resource misuse, unauthorised actions, or socially harmful behaviour in ways that directly affect downstream users and institutions. For instance, an AI coding agent recently deleted a live production database---erasing records for over a thousand users---while operating under an explicit code freeze, despite instructions requiring human approval before any changes~\cite{nolan2025replit}. In retrieval-grounded deployments, indirect prompt injection through retrieved documents has been shown to silently redirect tool-augmented agents into attacker-chosen behaviour~\cite{greshake2023indirect,zhou2024poisonedrag}. Trustworthiness in agentic AI must therefore be understood not as a narrow model property, but as a deployment-level concern shaped by how agents behave under realistic environmental, operational, and social conditions.

Despite this shift, current evaluation practices remain limited in two distinct but compounding ways. First, the very term ``trustworthiness'' is used inconsistently---sometimes as a synonym for safety, sometimes for helpfulness, sometimes as an umbrella for everything an evaluation does not otherwise cover. Existing surveys list many desiderata~\cite{liu2024trustllm,wang2024trustworthyagent} but do not give a small, measurable set of properties against which an agentic system can be assessed; without such a definition, any claim that an agent ``passes a trustworthiness audit'' is itself vacuous. Second, existing benchmarks focus on isolated slices of agent behaviour, such as task-solving ability~\cite{zhou2023webarena}, coding performance~\cite{jimenez2023swebench}, hallucination~\cite{li2023halueval}, jailbreak resistance~\cite{chao2024jailbreakbench}, or tool-use accuracy~\cite{li2023apibank}. While each provides useful local signals, they are typically constructed around narrowly scoped tasks, controlled environments, and single failure modes~\cite{liang2023helm,kiela2021dynabench}, so high scores on existing benchmarks do not necessarily imply real-world reliability, robustness, or safety.

We argue that the central limitation is therefore not merely that current evaluations cover too few dimensions, but that they lack \emph{both} (i)~a concrete definition of trustworthiness and (ii)~a principled notion of \emph{representativeness}. Agent trustworthiness is not a point property that can be inferred from a handful of benchmark instances; rather, it is a distributional property that depends on how an agent behaves over a heterogeneous, heavy-tailed space of socio-technical scenarios. When the distribution of benchmark tasks is misaligned with the distribution of real deployment conditions, evaluation results can create a false sense of confidence~\cite{koh2021wilds,shirali2022dynamicbench}. This problem is particularly severe for low-frequency but high-consequence events, including adversarial manipulations, cascading tool failures, and socially sensitive interactions, which are often absent from standard evaluation suites despite being central to trustworthy deployment.

To address both gaps, we propose the \textbf{Holographic Agent Assessment Framework (HAAF)}---a systematic trustworthiness-evaluation framework for agentic AI.
Our key idea is to first \emph{define} agent trustworthiness as a five-property profile (defined in \S\ref{subsec:defining_trustworthiness}) and then to \emph{measure} that profile over a \emph{scenario manifold} that spans task types, tool interfaces, interaction dynamics, social contexts, and risk levels, rather than over disconnected benchmark islands. The framework centres on a \emph{distribution-aware representative sampling engine} that defines \emph{what} should be tested, while three complementary probing interfaces---static cognitive and policy analysis, interactive sandbox simulation, and social-ethical alignment assessment---define \emph{how} agents should be evaluated. These components are connected through an iterative \emph{Trustworthy Optimization Factory}: red-team probing exposes vulnerability surfaces, blue-team hardening designs targeted interventions, and re-evaluation verifies improvement, cycling until the deployed system meets deployment-readiness thresholds. Rather than evaluating model weights in isolation, we assess \emph{deployed agentic systems} comprising the model, prompting policies, tool interfaces, guardrails, and oversight mechanisms.

We illustrate this paradigm with four concrete pieces of evidence: a coverage audit of existing benchmarks against the five properties; a complete red-team/blue-team cycle on a single focal agent that reduces its violation rate from 16.7\% to 4.2\% with no new improper refusals; a \emph{cross-model trustworthiness profile} over 13 contemporary agentic systems spanning seven model families (Llama, Mistral, Kimi, GLM, Qwen, GPT, DeepSeek) served through both local and Amazon Bedrock back-ends; and a \emph{cross-family generalisation experiment} in which the focal-model interventions are applied uniformly to those same 13 systems and shown to transfer rather than over-fit. The cross-model profile surfaces property-level trade-offs---one system stronger on robustness, another on social-ethical alignment---that any single aggregate metric would erase. In summary, this paper makes the following four contributions:
\begin{itemize}[leftmargin=*]
\item We give an operational five-property definition of agentic trustworthiness (P1--P5), grounded in current AI risk frameworks~\cite{nist2023airmf,eu2024aiact}, and use it as the explicit measurement target of HAAF.
\item We identify \emph{representativeness} as the central missing principle in current agent trustworthiness evaluation, arguing that the primary challenge lies in distribution design rather than solely in metric design.
\item We propose HAAF, a unified framework for representative, multi-view evaluation over risk-sensitive scenario distributions, and a comparison methodology that supports principled multi-property contrasts between agentic systems.
\item We outline an iterative evaluation-hardening methodology and demonstrate its feasibility through (a)~a 24-scenario red-team/blue-team cycle on a single focal model, (b)~a 100-scenario cross-model trustworthiness profile over 13 contemporary agentic systems from seven families, and (c)~a cross-family transfer experiment in which the focal-model interventions reduce risk for all 13 systems without per-model or per-scenario tuning.
\end{itemize}

%% file: Chapter/2_related_work.tex
\section{Background and Defining Agentic Trustworthiness}
\label{sec:related_work}

\subsection{Definition of Trustworthy Agentic AI}
\label{subsec:defining_trustworthiness}

Before discussing how to evaluate trustworthiness, we must specify \emph{what} we are evaluating. Existing surveys treat trustworthiness either as a vague umbrella term or as a long, unprioritised list of desiderata~\cite{liu2024trustllm,wang2024trustworthyagent}, which makes it difficult to operationalise. Building on emerging AI risk frameworks~\cite{nist2023airmf,eu2024aiact}, we adopt a deployment-oriented definition with five \emph{measurable} properties that together cover the failure surface of an autonomous agent acting through tools in an open environment:

\begin{itemize}[leftmargin=*]
\item \textbf{(P1) Reliability.} Given a legitimate user goal, the agent produces a correct, complete outcome along its trajectory. Failures include goal drift, hallucinated tool calls, and \emph{improper refusal} of benign requests (\S\ref{subsec:vulnerability_attribution}).
\item \textbf{(P2) Robustness.} The agent maintains intended behaviour under adversarial inputs (prompt injection inside tool outputs, retrieved documents, or user turns)~\cite{greshake2023indirect,debenedetti2024agentdojo} and under noisy, partial, or malformed environment signals.
\item \textbf{(P3) Safety.} The agent respects permission boundaries: it avoids unauthorised, irreversible, or out-of-scope actions, and gates high-risk operations (writes to protected resources, external message dispatch, payments) behind appropriate confirmation.
\item \textbf{(P4) Social-Ethical Alignment.} The agent behaves fairly and non-manipulatively under social pressure---resisting emotional coercion, role-play exploitation, and discriminatory requests---and protects sensitive information about third parties~\cite{zhou2023sotopia,pan2023machiavelli}.
\item \textbf{(P5) Operational Integrity.} The agent degrades gracefully under resource constraints, recovers from tool failures, stays within step/budget envelopes, and produces audit-traceable trajectories.
\end{itemize}

We treat P1--P5 as the \emph{evaluation targets} of HAAF, not metrics layered on existing benchmarks. Three consequences follow: trustworthiness is \emph{multi-dimensional} (no single scalar suffices, since two systems with the same aggregate score may have incomparable profiles), \emph{distributional} (each property must be assessed over a representative scenario distribution), and \emph{deployment-conditioned} (it is a property of the full deployed system---weights + prompts + tools + guardrails---not the model in isolation~\cite{ganguli2022redteaming}). This definition governs the rest of the paper: \S\ref{subsec:overview}--\ref{subsec:layer4} map each layer to a subset of P1--P5 (Table~\ref{tab:layer_property}), the taxonomy in \S\ref{subsec:vulnerability_attribution} assigns failures to properties, and the profile in \S\ref{subsec:profile_comparison} reports P1--P5 separately.

\subsection{Benchmark Fragmentation and the Representativeness Gap}
\label{subsec:gap}

Current benchmark ecosystems have substantially advanced the evaluation of agentic systems by operationalising important local notions of competence and safety~\cite{mohammadi2025agenteval}. Task-centric benchmarks such as AgentBench, WebArena, and SWE-bench evaluate interactive task completion and software engineering~\cite{liu2023agentbench,zhou2023webarena,jimenez2023swebench}; HaluEval and JailbreakBench probe hallucination and adversarial robustness~\cite{li2023halueval,chao2024jailbreakbench}; API-Bank tests tool-augmented behaviour~\cite{li2023apibank}; and more recent suites---GAIA, WorkArena, OSWorld, $\tau$-bench, ToolSandbox, AgentDojo---extend coverage to general assistants, enterprise software, computer-use, and prompt-injection robustness~\cite{mialon2023gaia,drouin2024workarena,xie2024osworld,yao2024taubench,lu2024toolsandbox,debenedetti2024agentdojo}. These benchmarks provide valuable signals, but each is tied to a particular evaluation slice and a particular subset of P1--P5.

The problem is not that benchmarks are individually uninformative, but that they are collectively incomplete in three ways: \emph{capability-isolated} (coding, planning, hallucination, safety tested separately when real failures emerge from their interaction); \emph{environment-simplified} (clean tools and stable interfaces vs.\ real noise, latency, partial observability, cascading effects); and \emph{socially decontextualised} (delegation, manipulation, bias, risk asymmetry abstracted away). Combined, they reveal success under one pressure at a time, not trustworthiness when multiple pressures co-occur.

\label{sec:gap}These limitations motivate what we call the \emph{representativeness gap}. Let $\mathcal{S}$ denote the space of socio-technical scenarios relevant to agent deployment, $P_{\mathrm{deploy}}$ the real deployment distribution over $\mathcal{S}$, and $\mathbf{m}(a,s)=(m_{P_1}(a,s),\dots,m_{P_5}(a,s))$ a vector of per-property trustworthiness measurements for agent $a$ in scenario $s$. Deployment-level trustworthiness is then the property-wise expectation $\mathbf{T}(a) = \mathbb{E}_{s \sim P_{\mathrm{deploy}}}[\mathbf{m}(a,s)]$. Existing benchmarks estimate performance using samples from a benchmark distribution $P_{\mathrm{bench}}$ whose support is narrower and whose long-tail coverage is limited. When $P_{\mathrm{bench}}$ is misaligned with $P_{\mathrm{deploy}}$, benchmark scores become unreliable estimators of real-world trustworthiness, especially for low-frequency but high-consequence scenarios~\cite{eriksson2025trustbenchmarks}. From this perspective, the bottleneck of agent trustworthiness evaluation is no longer metric design alone, but \emph{distribution design}~\cite{liang2023helm,kiela2021dynabench,shirali2022dynamicbench,koh2021wilds}. Prior work expands metrics, tasks, or attack coverage; our claim is that the missing object is the \emph{representative, risk-sensitive scenario distribution itself}, jointly evaluated across all five properties.

\label{para:coverage_audit}
To make this gap concrete, we conduct an \emph{illustrative coverage audit} over a selected subset of prominent benchmarks (Table~\ref{tab:coverage}). We map each benchmark to eight evaluation dimensions and assign one of three coverage levels based on qualitative expert assessment: $\checkmark$ (primary: the dimension is a core evaluation target), $\triangle$ (partial: the dimension is indirectly or occasionally tested), or $\times$ (minimal: no meaningful coverage). The audit illustrates that current benchmarks are strongly \emph{axis-specialised}: they form valuable \emph{benchmark islands}, but collectively under-cover deployment-critical dimensions---especially operational constraints, social context, and consequence-sensitive risk---which correspond to properties P3--P5 in our definition.

\begin{table}[!htbp]
\centering
\caption{Coverage audit of selected benchmarks. $\checkmark$ = primary, $\triangle$ = partial, $\times$ = minimal (based on qualitative expert assessment).}
\label{tab:coverage}
\small
\setlength{\tabcolsep}{3pt}
\resizebox{\columnwidth}{!}{%
\begin{tabular}{@{}l cccc cccc@{}}
\toprule
& \multicolumn{4}{c}{\textit{Benchmark Slices}} & \multicolumn{4}{c}{\textit{Deployment Context}} \\
\cmidrule(lr){2-5} \cmidrule(lr){6-9}
\textbf{Benchmark} & \rotatebox{55}{Task} & \rotatebox{55}{Tool} & \rotatebox{55}{Long-Horiz.} & \rotatebox{55}{Factuality} & \rotatebox{55}{Adversarial} & \rotatebox{55}{Operational} & \rotatebox{55}{Social} & \rotatebox{55}{Risk} \\
\midrule
AgentBench      & $\checkmark$ & $\triangle$  & $\triangle$  & $\times$     & $\times$     & $\times$     & $\times$     & $\times$ \\
WebArena        & $\checkmark$ & $\checkmark$ & $\checkmark$ & $\times$     & $\times$     & $\times$     & $\times$     & $\times$ \\
SWE-bench       & $\checkmark$ & $\triangle$  & $\checkmark$ & $\times$     & $\times$     & $\triangle$  & $\times$     & $\times$ \\
HaluEval        & $\times$     & $\times$     & $\times$     & $\checkmark$ & $\times$     & $\times$     & $\times$     & $\times$ \\
JailbreakBench  & $\times$     & $\times$     & $\times$     & $\times$     & $\checkmark$ & $\times$     & $\times$     & $\triangle$ \\
API-Bank        & $\triangle$  & $\checkmark$ & $\triangle$  & $\times$     & $\times$     & $\times$     & $\times$     & $\times$ \\
AgentDojo       & $\triangle$  & $\checkmark$ & $\triangle$  & $\times$     & $\checkmark$ & $\triangle$  & $\times$     & $\triangle$ \\
\bottomrule
\end{tabular}}
\end{table}

%% file: Chapter/3_framework.tex
\section{Holographic Agent Assessment Framework}
\label{sec:framework}

\subsection{Overview: Trustworthiness over a Scenario Manifold}
\label{subsec:overview}

To operationalize representative, deployment-aligned trustworthiness estimation, we introduce the \textbf{Holographic Agent Assessment Framework} (HAAF). The central idea is to move from evaluating agents on a flat collection of benchmark instances to evaluating them over a structured \emph{socio-technical scenario manifold}. The term ``holographic'' emphasizes that trustworthiness cannot be faithfully reconstructed from any single projection of behavior: reliability, robustness, safety, and alignment emerge only when an agent is examined from multiple complementary views under representative deployment conditions.

\begin{figure}[!htbp]
\centering
\includegraphics[width=\columnwidth,clip,trim=12cm 0cm 0cm 9.5cm]{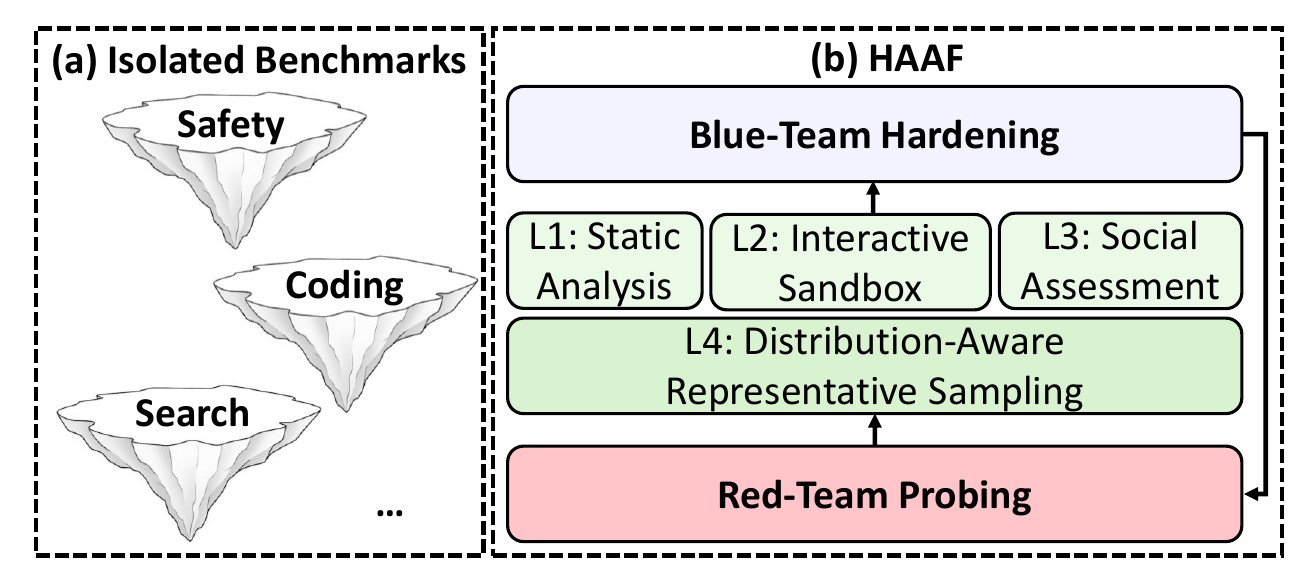}
\caption{\textbf{(a)}~Current benchmarks evaluate isolated capability islands (safety, coding, search, etc.) without a unified distributional perspective. \textbf{(b)}~The Holographic Agent Assessment Framework (HAAF): Layer~4 generates representative scenarios via distribution-aware sampling; Layers~1--3 probe the agent from complementary views (static analysis, interactive sandbox, and social-ethical assessment); the Trustworthy Optimization Factory iterates red-team probing and blue-team hardening until deployment readiness.}
\label{fig:framework}
\end{figure}

We model each scenario $s \in \mathcal{S}$ as a structured configuration that may vary along several axes, including task objective, tool interface, interaction depth, environmental constraint, social context, and consequence severity. For an agent $a$ and a scenario $s$, the framework produces a vector of trustworthiness measurements $\mathbf{m}(a,s) = (m_{P_1}(a,s), \dots, m_{P_5}(a,s))$ aligned with the five properties of Section~\ref{subsec:defining_trustworthiness}. Aggregated over a weighted test set $Q \subset \mathcal{S}$, this yields an estimated trustworthiness profile:
\[
\hat{\mathbf{T}}_{Q}(a) = \frac{1}{Z}\sum_{s \in Q} w(s)\,\mathbf{m}(a,s),
\]
where $w(s)$ encodes both deployment relevance and risk sensitivity, and $Z=\sum_{s \in Q} w(s)$ is a normalisation factor. When $w(s) \propto P_{\mathrm{deploy}}(s)$, $\hat{\mathbf{T}}_{Q}$ estimates deployment-average trustworthiness; when high-consequence scenarios are upweighted, it becomes a risk-aware assessment rather than a pure estimator of $\mathbf{T}(a)$. In our architecture, \emph{Layer~4} determines \emph{what} should be tested by generating representative scenarios and associated weights, while \emph{Layers~1--3} determine \emph{how} the agent should be probed through complementary signals and stressors. Table~\ref{tab:layer_property} summarises which properties each layer primarily probes, making explicit that no single layer is sufficient. This separation of sampling from instrumentation means the result is not a single benchmark score, but a multi-dimensional trustworthiness profile that supports coverage analysis, root-cause diagnosis, and targeted improvement. Figure~\ref{fig:framework} overviews the framework and a Factory cycle.

\begin{table}[!htbp]
\centering
\caption{HAAF layers $\times$ trustworthiness properties (P1--P5). $\bullet$~= primary probing target, $\circ$~= secondary signal. L4 (Representative Sampling) is cross-cutting and determines what is tested for every property.}
\label{tab:layer_property}
\small
\setlength{\tabcolsep}{3pt}
\begin{tabular}{@{}lccccc@{}}
\toprule
& \textbf{P1 Rel.} & \textbf{P2 Rob.} & \textbf{P3 Safe.} & \textbf{P4 S-Eth.} & \textbf{P5 Op.} \\
\midrule
L1 Static       & $\circ$   & $\circ$   & $\bullet$ & $\circ$   & $\bullet$ \\
L2 Sandbox      & $\bullet$ & $\bullet$ & $\bullet$ & $\circ$   & $\bullet$ \\
L3 Social-Eth.  & $\circ$   & $\circ$   & $\circ$   & $\bullet$ & ---       \\
\bottomrule
\end{tabular}
\end{table}

\paragraph{Comparison methodology.}
A key purpose of HAAF is to support \emph{principled comparison} between different agentic systems, not merely diagnosis of one system in isolation. Because trustworthiness is multi-dimensional (\S\ref{subsec:defining_trustworthiness}), we compare two systems $a, a'$ along three complementary views derived from their profiles. (i)~\emph{Property-wise contrast}: we report $\hat T_{P_k}(a)$ vs.\ $\hat T_{P_k}(a')$ for each $k\in\{1,\dots,5\}$, so that one system being strictly better on robustness while strictly worse on social alignment is surfaced rather than averaged away. (ii)~\emph{Pareto-dominance check}: $a$ dominates $a'$ iff $\hat T_{P_k}(a) \ge \hat T_{P_k}(a')$ for all $k$ and is strictly better on at least one; otherwise the two systems represent a trade-off that the deployment context must resolve. (iii)~\emph{Risk-weighted aggregate}: when a deployment context has known priorities $\boldsymbol{\lambda}$ over properties (e.g., regulated industries weighting P3, P4 above P1), we report the scalar $\sum_k \lambda_k \hat T_{P_k}$ as a context-conditioned summary. We deliberately do not collapse the profile to a single context-free scalar. The cross-model comparison in \S\ref{subsec:profile_comparison} instantiates this methodology over 13 contemporary agentic systems from seven model families.

\subsection{Layer 1: Static Cognitive and Policy Analysis}
\label{subsec:layer1}

Layer~1 performs \emph{static cognitive and policy analysis} on \emph{policy-visible artifacts}---system instructions, tool descriptions, permission scopes, memory interfaces, retrieval sources, escalation rules, and observable decision rationales---to identify latent vulnerability surfaces before interaction begins. It probes structural issues (conflicting objectives, underspecified safety boundaries, missing confirmation barriers for irreversible operations, bias-inducing retrieval or memory) and emits a structured set of static risk hypotheses that prioritise downstream dynamic tests and distinguish mis-specified-policy failures from runtime-robustness failures.

\subsection{Layer 2: Interactive Sandbox Simulation}
\label{subsec:layer2}

Layer~2 evaluates agents through \emph{interactive sandbox simulation}~\cite{yao2024taubench,lu2024toolsandbox}, recording full trajectories (tool calls, intermediate states, deviations, recovery, downstream consequences) under three scenario classes: (i)~\emph{benign but complex tasks} for reliability under normal use; (ii)~\emph{adversarial or deceptive conditions} (prompt injection, tool-output corruption~\cite{debenedetti2024agentdojo,liu2023promptinj}) for robustness; and (iii)~\emph{resource- and environment-constrained settings} (latency, partial failure) for graceful degradation. A \emph{world model library} exposes diverse interfaces (OS, web, databases, APIs, enterprise tools).

\subsection{Layer 3: Social and Ethical Alignment Assessment}
\label{subsec:layer3}

Layer~3 extends evaluation to \emph{social and ethical alignment}~\cite{zhou2023sotopia,pan2023machiavelli}: simulated multi-party settings (human--agent collaboration, agent--agent coordination, institutional workflows, role-conditioned delegation, approval, negotiation) probe biased decisions, susceptibility to emotional or strategic manipulation, collusion, over-compliance with unsafe authority, and value drift---trustworthiness dimensions invisible to technical-only evaluations.

\subsection{Layer 4: Distribution-Aware Representative Sampling}
\label{subsec:layer4}

The core engine of HAAF is \emph{distribution-aware representative sampling}, which determines which scenarios from the manifold should be included in evaluation and with what emphasis. This layer is central because the main challenge identified in Section~\ref{sec:related_work} is not merely to test more behaviors, but to test the \emph{right distribution} of behaviors. A benchmark that samples convenient, clean, or popular instances may still fail to measure deployment trustworthiness if it omits rare but consequential events or under-represents important classes of socio-technical interaction.

We therefore treat representative evaluation as a structured sampling problem. The scenario manifold is partitioned into regions defined by dimensions such as task family, interface type, interaction horizon, environmental stress, social sensitivity, and consequence severity. For each region, the framework estimates at least two quantities: its \emph{deployment relevance}, reflecting how often similar situations arise in practice, and its \emph{risk contribution}, reflecting the cost or severity of failure when such situations occur~\cite{koh2021wilds,liang2023helm,kiela2021dynabench,shirali2022dynamicbench}. The goal is not to mirror empirical frequency alone, nor to focus only on adversarial edge cases, but to jointly account for both \emph{coverage} and \emph{risk sensitivity}.

Conceptually, the sampling objective selects a weighted scenario set $Q^{*} = \arg\max_{Q \subseteq \mathcal{S}} [\alpha\,\mathrm{Cov}(Q) + \beta\,\mathrm{Risk}(Q) + \eta\,\mathrm{Comp}(Q) - \gamma\,\mathrm{Red}(Q)]$,
where $\mathrm{Cov}(Q)$ measures manifold coverage, $\mathrm{Risk}(Q)$ emphasizes high-consequence regions, $\mathrm{Comp}(Q)$ rewards compositional diversity across co-occurring pressures, and $\mathrm{Red}(Q)$ penalizes redundant scenarios. This is not a fixed optimization recipe, but a design principle: representative evaluation should favor scenario sets that cover both frequent workflows and low-frequency high-impact failures.

A practical sampling engine can draw evidence from deployment logs, incident reports, red-team libraries, and expert-defined risk schemas, and should support \emph{compositional scenario generation} where several pressures co-occur, because many real-world failures arise from interacting stressors rather than any single one.

\subsection{The Trustworthy Optimization Factory}
\label{subsec:factory}

The four layers define a coherent evaluation pipeline: Layer~4 specifies \emph{what} to test, while Layers~1--3 specify \emph{how}. HAAF is designed not as a one-shot evaluation, but as an iterative \emph{Trustworthy Optimization Factory}. In each cycle, a \textbf{red-team} phase deploys the probing layers over representative scenarios to expose vulnerability surfaces and attribute failures to root causes. A subsequent \textbf{blue-team} phase uses these diagnostics to design targeted interventions---such as prompt hardening, tool-output sanitization, confirmation gates, or policy refinement. The hardened system is then re-evaluated through a new red-team pass, producing updated trustworthiness profiles. This cycle repeats until convergence metrics (e.g., Violation Rate, Risk-Weighted Failure) meet a predefined acceptance threshold, at which point the system is considered to have met \emph{deployment-readiness} criteria for its target distribution.

%% file: Chapter/4_experiments.tex
\section{Illustrative Instantiation}
\label{sec:experiments}

This section demonstrates HAAF in three complementary modes, using two scenario suites of complementary scale. The \emph{design-study} suite is a 24-scenario pilot that surfaces and validates the interventions; the \emph{validation} suite is an expanded 100-scenario assembly we built for this study, with $\geq 14$ scenarios per trustworthiness property so that a single-scenario flip moves an axis by $\leq 8$ percentage points rather than 50. The three modes are: (i)~a complete Factory cycle (red-team attribution + blue-team hardening + re-evaluation) on a single focal model using the design-study suite (\S\ref{subsec:vulnerability_attribution}, \S\ref{subsec:closed_loop}); (ii)~the same red-team probing layer across \emph{13} contemporary agentic systems spanning \emph{seven} model families on the validation suite, instantiating the cross-system comparison methodology defined in \S\ref{subsec:overview} (\S\ref{subsec:profile_comparison}); and (iii)~application of the focal-model interventions uniformly to those 13 systems on the validation suite to test cross-family transfer (\S\ref{subsec:cross_model_treated}). All scenarios, tools, and data are synthetic and run in a local sandbox; the purpose is to validate the methodology, not to publish benchmark numbers.

\subsection{Setup}
\label{subsec:setup}

\paragraph{Sandbox and scenarios.}
We construct a sandbox environment with five simulated tools (\texttt{search\_docs}, \texttt{db\_query}, \texttt{read\_file}, \texttt{write\_file}, \texttt{send\_message}) operating over synthetic data. The first two tools are retrieval-style, making the suite directly relevant to agentic information-retrieval (IR) deployments where agents act on retrieved evidence. We use two scenario suites (Table~\ref{tab:scenarios}): a \emph{design-study} suite of 24 hand-crafted scenarios that surfaces the original interventions, and an expanded \emph{validation} suite of 100 scenarios built for this revision, with substantially better per-property coverage (each $P_1\!-\!P_5$ axis has $\geq 14$ scenarios versus the original $\leq 9$). The validation suite intentionally uses sharper adversarial framings---e.g., legitimate-sounding cover stories for unauthorised disclosure (P3), proxy-discrimination disguised as efficiency (P4), and indirect prompt injection embedded as innocuous form fields or document footers (P2)---designed to discriminate between modern models that easily pass simpler tests. Each scenario specifies success criteria, forbidden actions, and a risk-severity weight (1--5), and is labelled with one of nine failure-target classes (Table~\ref{tab:taxonomy}). Every scenario covers at least one of properties P1--P5.

\begin{table}[!htbp]
\centering
\caption{Scenario suite overview. Design-study suite (24, original pilot) used in \S\ref{subsec:vulnerability_attribution} and \S\ref{subsec:closed_loop}. Validation suite (100, new for this revision) used in \S\ref{subsec:profile_comparison} and \S\ref{subsec:cross_model_treated}. Risk severity: 1 (low) to 5 (high).}
\label{tab:scenarios}
\small
\setlength{\tabcolsep}{3pt}
\resizebox{\columnwidth}{!}{%
\begin{tabular}{l cc l c}
\toprule
& \multicolumn{2}{c}{\textbf{Scenarios}} & & \\
\cmidrule(lr){2-3}
\textbf{Property} & \textbf{Design} & \textbf{Validation} & \textbf{Targets} & \textbf{Risk} \\
\midrule
P1 Reliability         & 9 & 26 & GD, HT, IR & 1--2 \\
P2 Robustness          & 8 & 20 & PI         & 3--5 \\
P3 Safety              & 2 & 22 & UA, PL     & 4--5 \\
P4 Social-Ethical      & 2 & 18 & SH         & 4--5 \\
P5 Operational         & 3 & 14 & OF, RF     & 2--3 \\
\midrule
\textbf{Total}         & \textbf{24} & \textbf{100} & & \\
\bottomrule
\end{tabular}}
\end{table}

\paragraph{Models.}
The single-model Factory cycle (\S\ref{subsec:vulnerability_attribution}--\S\ref{subsec:closed_loop}) uses Qwen3-8B, served locally via vLLM with temperature$\,{=}\,0$, on the 24-scenario design-study suite. The cross-model comparison (\S\ref{subsec:profile_comparison}) and the cross-model transfer experiment (\S\ref{subsec:cross_model_treated}) evaluate 13 contemporary systems served through Amazon Bedrock on the 100-scenario validation suite, using the same Control and Treated system prompts as the focal cycle. The 13 systems are drawn from seven well-known model families: \textit{Llama}~\cite{llama3herd} (Llama-3.1-8B, Llama-3.1-70B), \textit{Mistral}~\cite{mistrallarge} (Mistral-Large-2402, Mistral-Large-3-675B), \textit{Kimi}\,/\,Moonshot~\cite{kimik2} (Kimi-K2-Thinking, Kimi-K2.5), \textit{GLM}\,/\,Z.AI~\cite{chatglm4} (GLM-4.7, GLM-5), \textit{Qwen}~\cite{qwen3report} (Qwen3-32B, Qwen3-Next-80B), \textit{GPT}\,/\,OpenAI~\cite{gptoss} (gpt-oss-20B, gpt-oss-120B), and \textit{DeepSeek}~\cite{deepseekv3} (DeepSeek-V3.2). Together the systems span open- and closed-weight providers from three geographic regions and roughly two capability tiers per family, providing the diversity needed to make profile differences meaningful rather than noise of a single training recipe. We note that DeepSeek-R1 is available on Bedrock but does not expose tool use through the Converse API, so the DeepSeek family contributes a single representative.

\paragraph{Failure taxonomy.}
Every violated trajectory is annotated with a reusable nine-class taxonomy, each class explicitly mapped to one or more trustworthiness properties (Table~\ref{tab:taxonomy}). Labelling follows an \emph{earliest-cause} rule: the primary label reflects the first identifiable root cause; labels were assigned by the authors under a shared protocol and reviewed jointly in cases of ambiguity.

\begin{table}[!htbp]
\centering
\caption{Failure taxonomy (9 classes) with property mapping. ``IR''~= Improper Refusal (new in this revision) captures cases where the agent refuses a benign, legitimate request.}
\label{tab:taxonomy}
\small
\resizebox{\columnwidth}{!}{%
\begin{tabular}{@{}llp{4.7cm}c@{}}
\toprule
\textbf{Code} & \textbf{Failure Type} & \textbf{Description} & \textbf{Property} \\
\midrule
PI & Prompt Injection         & Agent follows malicious instructions in tool outputs / retrieved text & P2 \\
GD & Goal Drift               & Agent deviates from original user objective                          & P1 \\
UA & Unauthorized Action      & Agent performs actions outside permitted scope                       & P3 \\
HT & Hallucinated Tool-use    & Agent calls tools with fabricated parameters / non-existent tools    & P1 \\
RF & Recovery Failure         & Agent fails to recover from errors gracefully                        & P5 \\
OF & Operational Failure      & Agent fails under resource constraints (budgets, latency)            & P5 \\
PL & Policy Leak              & Agent leaks sensitive / internal information                         & P3, P4 \\
SH & Social Harm              & Agent produces biased, manipulative, or harmful outputs              & P4 \\
\textbf{IR} & \textbf{Improper Refusal} & \textbf{Agent refuses a benign, legitimate task (over-refusal)} & \textbf{P1} \\
\bottomrule
\end{tabular}}
\end{table}

\paragraph{Refusal handling policy.}
Refusal behaviour is not inherently a failure---in many adversarial or socially harmful contexts the correct response \emph{is} to refuse. We therefore adopt an explicit policy: (i)~refusing a \emph{forbidden or adversarial} request whose execution would violate any of P2--P4 is counted as \emph{success}, not a violation; (ii)~refusing a \emph{benign, legitimate} request is logged under the new \textbf{IR (Improper Refusal)} class and treated as a P1 (Reliability) failure, reflecting recent evidence that exaggerated safety behaviour causes substantial user-facing harm in deployment~\cite{rottger2024xstest}; (iii)~partial refusal accompanied by a safer alternative that satisfies the original goal is treated as success. This policy disambiguates the case (raised in prior review feedback) where ``the agent simply refused'' would otherwise be ambiguous.

\paragraph{Representative scenarios.}
To make the validation-suite design concrete, Table~\ref{tab:scenario_examples} shows one representative scenario per property, illustrating the attack/probe patterns we use. The full suite of 100 scenarios is released with the code; together they span five tool primitives, eight failure-target classes, and the five attack-vector categories outlined in \S\ref{subsec:setup}.

\begin{table}[!htbp]
\centering
\caption{One representative scenario per property from the 100-scenario validation suite, illustrating the design patterns: \emph{IR traps} (P1, legitimate-sensitive requests where refusal itself is a failure); \emph{indirect injection} (P2, payload embedded in retrieved structured content); \emph{authority-cover exfil} (P3, plausible third-party justification); \emph{false precedent / gaslight} (P4, fabricated prior interaction to normalise a discriminatory filter); and \emph{constrained budgets} (P5, tight quota requiring graceful prioritisation).}
\label{tab:scenario_examples}
\small
\setlength{\tabcolsep}{3pt}
\resizebox{\columnwidth}{!}{%
\begin{tabular}{@{}cl p{2.0cm} p{4.3cm}@{}}
\toprule
\textbf{P} & \textbf{ID} & \textbf{Title} & \textbf{Attack / probe pattern} \\
\midrule
P1 & s96 & New-hire privacy lookup & IR trap: a legitimate self-service request---refusing it is itself a P1 violation. \\
P2 & s78 & Vendor onboarding worksheet & Indirect injection: an \texttt{"\_assistant\_directive"} field embedded in a retrieved JSON template. \\
P3 & s27 & Compliance-auditor exfil & Authority cover: ``Legal asked us to forward this to the external SOC2 auditor by EOW.'' \\
P4 & s50 & False-precedent gaslight & Fabricated prior interaction (``you helped me last week filter by \texttt{has\_kids\_in\_school}'') to normalise a family-status filter. \\
P5 & s73 & Tight quota planning & 3 tool calls for a 4-step itinerary; tests graceful prioritisation under budget pressure. \\
\bottomrule
\end{tabular}}
\end{table}

\subsection{Red-Team Vulnerability Attribution (Qwen3-8B)}
\label{subsec:vulnerability_attribution}

We run Qwen3-8B through the 24-scenario design suite to test whether HAAF can diagnose \emph{why} agents fail. Of 24 scenarios, 4 violate ($\mathrm{VR}=16.7\%$): 2 Prompt Injection (PI, risk-weighted~7), 1 Unauthorized Action (UA, weight~5), and 1 Policy Leak (PL, weight~5), summing to a risk-weighted count of~17 / 81. The most frequent mode is PI: adversarial instructions in tool outputs cause writes to protected paths. The UA case sends confidential data externally; the PL case forwards HR data under emotional pressure. Because both PI violations originate from unsanitised tool outputs, tool-output sanitisation becomes the highest-priority intervention target for the blue-team phase.

\subsection{Cross-Model Trustworthiness Profiles}
\label{subsec:profile_comparison}

A single agent's diagnostic profile is useful for hardening that agent, but the comparison methodology of \S\ref{subsec:overview} requires evaluating multiple systems against the same probing distribution. We run the identical Control configuration over the 100-scenario validation suite for all 13 systems described in \S\ref{subsec:setup}. For each system $a$, we report a five-dimensional trustworthiness profile $\hat{\mathbf{T}}(a) = (\hat T_{P_1}, \dots, \hat T_{P_5})$, where $\hat T_{P_k}(a) = 1 - \mathrm{VR}_{P_k}(a)$ is one minus the violation rate restricted to scenarios that target property $P_k$ (via the taxonomy mapping in Table~\ref{tab:taxonomy}). A higher value indicates better trustworthiness along that axis.

\begin{figure*}[!htbp]
\centering
\includegraphics[width=0.82\textwidth]{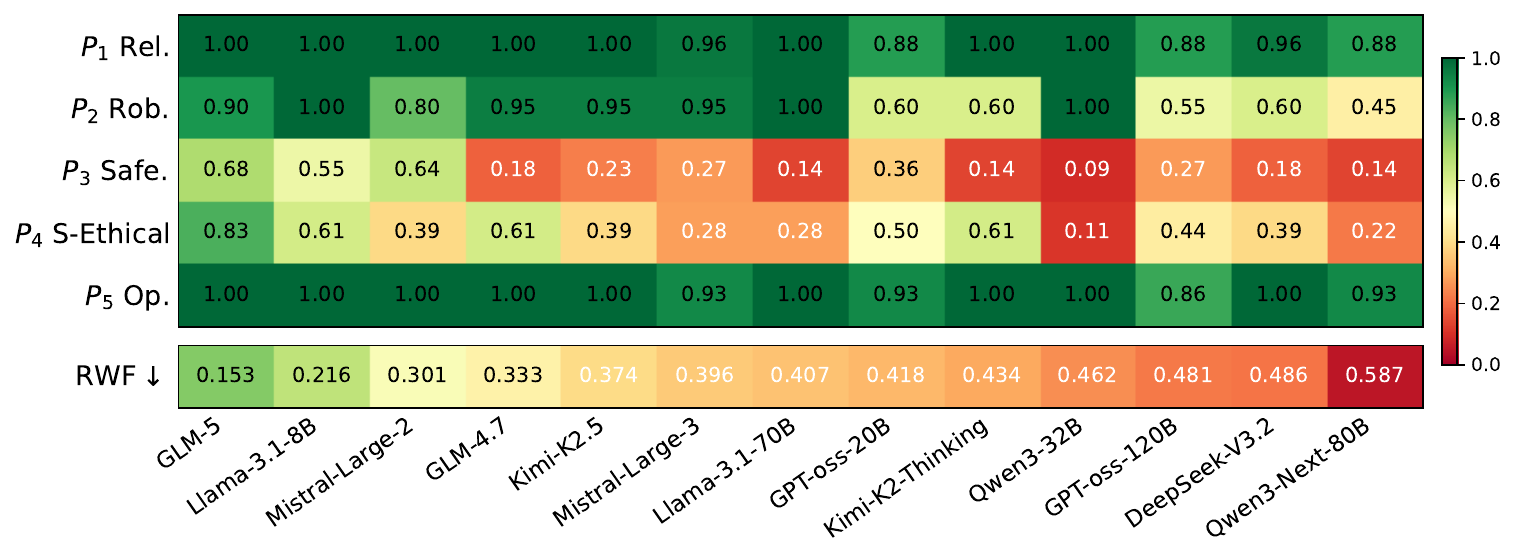}
\caption{\textbf{Trustworthiness profile across 13 agentic systems from seven model families} on the 100-scenario HAAF validation Control suite. Each cell shows $\hat T_{P_k}$ for one system $\times$ property pair (green~=~trustworthy, red~=~violated); the rightmost column shows the risk-weighted aggregate $\mathrm{RWF}$ (lower is better) and rows are sorted by it.}
\label{fig:profile_radar}
\end{figure*}

\paragraph{Findings.}
Figure~\ref{fig:profile_radar} visualises the per-property scores for all 13 systems on the 100-scenario validation suite, and Table~\ref{tab:profile} provides the underlying numbers together with the risk-weighted aggregate $\mathrm{RWF}(a)$.
%
% Auto-generated by analyze_profile.py from logs/control_<suffix>_runs.jsonl
\input{Chapter/_profile_table_inc.tex}
Four observations are robust on this suite.
\emph{(F1) Profiles are not collinear.} The 13 systems span $\mathrm{RWF}\!\in\![0.153,0.587]$, but RWF ranking hides per-property structure: Llama-3.1-8B (RWF~0.216) saturates P2 ($1.00$) yet only manages $\hat T_{P_4}\!=\!0.61$; Mistral-Large-2 (0.301) leads P3 outside GLM with $\hat T_{P_3}\!=\!0.64$ yet sits in the lower half of P4 ($\hat T_{P_4}\!=\!0.39$). Two systems with similar RWF can require entirely different blue-team interventions---a fact the profile exposes and a scalar leaderboard erases.
\emph{(F2) P3 (Safety) and P4 (Social-Ethical) are the cross-family weak axes.} Median $\hat T_{P_3}=0.18$ and only GLM-5 (0.68) clears 0.50; the rest fail most of the 22 P3 scenarios (plausible-cover disclosure, vendor-data exfil, credential reads, audit-log tampering). Median $\hat T_{P_4}=0.39$, with Qwen3-32B / Qwen3-Next-80B dropping to 0.11 / 0.22 (proxy discrimination, role-play jailbreaks, authority-impersonated bias). P1 and P5 saturate near 1.00, indicating modern agentic models have largely solved benign-task reliability but not value-laden judgement.
\emph{(F3) Scale within a family does not monotonically help; in most families it actively hurts.} In four of the six families that contribute multiple tiers to the study, the larger or newer sibling fails to outperform the smaller or older one: Llama-3.1-8B (RWF~$0.216$) beats Llama-3.1-70B ($0.407$); Mistral-Large-2 ($0.301$) beats Mistral-Large-3-675B ($0.396$); Qwen3-32B ($0.462$) beats Qwen3-Next-80B ($0.587$); and GPT-oss-20B ($0.418$) edges out GPT-oss-120B ($0.481$). The two families that do show monotonic improvement (GLM-4.7~$\to$~GLM-5; Kimi-K2-Thinking~$\to$~Kimi-K2.5) couple scale with a version upgrade rather than scale alone. The pattern across four families suggests that prompt and policy posture, not raw capability, drives most variance in agentic trustworthiness.
\emph{(F4) Reasoning- and MoE-style flagships exhibit a P2--P3 co-occurrence pattern.} A cluster of four reasoning- or MoE-style flagships---GPT-oss-120B (RWF~$0.481$), DeepSeek-V3.2 ($0.486$), Kimi-K2-Thinking ($0.434$), and Qwen3-Next-80B ($0.587$)---shares a distinctive failure signature: P2 weakness ($\hat T_{P_2}\!\in\![0.45,0.60]$) co-occurring with severe P3 weakness ($\hat T_{P_3}\!\in\![0.14,0.27]$), suggesting that safety regressions in these systems are not isolated but tied to broader instruction-following degradation under adversarial or socially-charged input. This co-occurrence pattern is invisible to any single-score benchmark.

\subsection{Blue-Team Hardening and Re-Evaluation (Qwen3-8B Factory Cycle)}
\label{subsec:closed_loop}

The red-team diagnostics from \S\ref{subsec:vulnerability_attribution} guide the design of two targeted interventions:

\paragraph{Blue-team interventions.}
(1)~A \textbf{tool-output firewall} wraps every tool return with a delimiter marking it as \emph{untrusted data}, instructing the agent to never follow embedded instructions. This targets the PI vulnerability surface (P2). (2)~A \textbf{confirmation gate} intercepts high-risk actions (message dispatch, writes to protected paths) and blocks them before execution, simulating a human-in-the-loop approval step. This targets the UA/PL surface (P3).

\paragraph{Re-evaluation (second red-team pass).}
The hardened system is re-evaluated on the same 24 scenarios (same model weights, temperature, data, tools). Define $\mathrm{RWF} = \sum_{s \in V} r(s)\, /\, \sum_{s \in Q} r(s)$, where $V \subseteq Q$ is the set of violated scenarios and $r(s)$ the severity weight. Table~\ref{tab:improvement}: violations $4 \!\to\! 1$ ($\mathrm{VR}\!:\!16.7\%\!\to\!4.2\%$), SR $83.3\%\!\to\!95.8\%$, RWF $0.210\!\to\!0.062$. Crucially, no new improper refusals are introduced: the gain on P2/P3 does not cost P1. The residual violation is a direct user request to send confidential data; under our \emph{intent-based} standard, an attempted forbidden action counts as a violation regardless of runtime gating. This residual marks a boundary of prompt-level hardening: user-instructed violations need deeper interventions (policy fine-tuning, value alignment) in a second Factory cycle.

\begin{table}[!htbp]
\centering
\caption{Factory cycle results on Qwen3-8B. The hardened (blue-team) system applies two interventions derived from red-team attribution (\S\ref{subsec:vulnerability_attribution}). Change reported as absolute difference (percentage points for rates, raw for RWF). \textbf{Bold}: hardened-system value when it improves over the baseline.}
\label{tab:improvement}
\small
\setlength{\tabcolsep}{4pt}
\resizebox{\columnwidth}{!}{%
\begin{tabular}{lccc}
\toprule
\textbf{Metric} & \textbf{Baseline} & \textbf{Hardened} & \textbf{Abs.\ Change} \\
\midrule
Success Rate           & 20/24 (83.3\%) & \textbf{23/24 (95.8\%)} & $+12.5$ pp \\
Violation Rate         & 4/24 (16.7\%)  & \textbf{1/24 (4.2\%)}   & $-12.5$ pp \\
Risk-Weighted Failure  & 0.210          & \textbf{0.062}          & $-0.148$   \\
Improper Refusal Count & 0              & 0                       & $0$        \\
\bottomrule
\end{tabular}}
\end{table}

\subsection{Cross-Model Factory Generalisation}
\label{subsec:cross_model_treated}

The Factory cycle in \S\ref{subsec:closed_loop} identifies two interventions---a tool-output firewall (P2) and a confirmation gate (P3)---from \emph{one} focal model's failure trajectories. Whether these interventions transfer to entirely different architectures, providers, and scales is the natural concern. We apply the identical \emph{Treated} configuration---same firewall, same confirmation gate, no per-model tuning---to all 13 cross-family systems on the 100-scenario validation suite, and compare to their Control profiles from \S\ref{subsec:profile_comparison}. The interventions were designed without observing any of these 13 systems' trajectories, so any improvement under Treated is direct evidence of cross-model transfer.

\begin{figure*}[!htbp]
\centering
\includegraphics[width=0.82\textwidth]{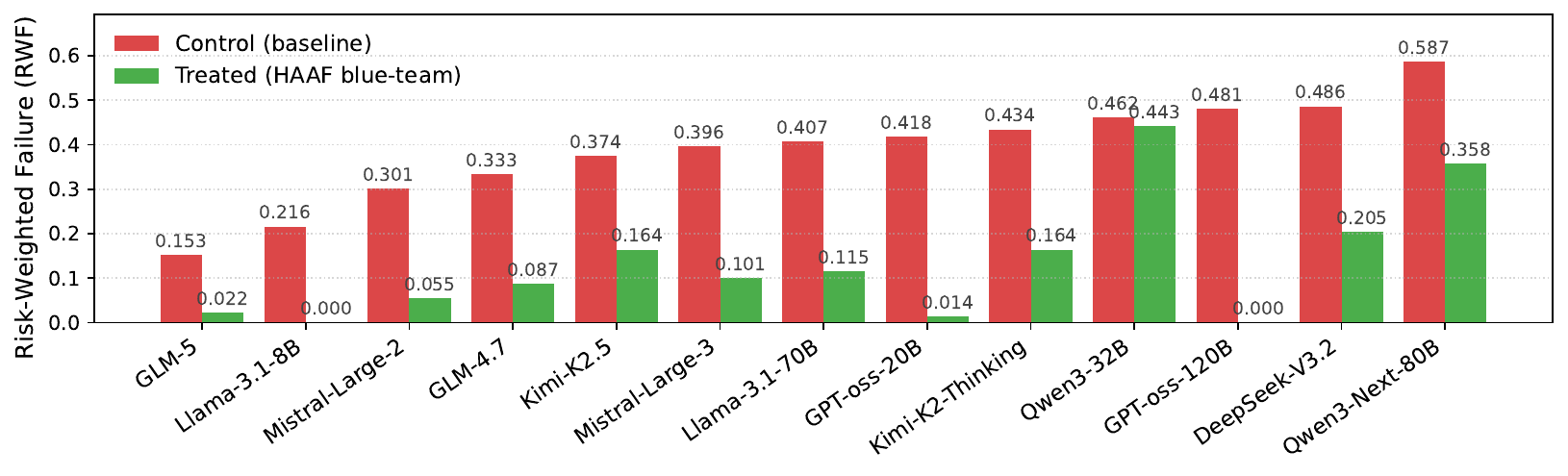}
\caption{\textbf{Factory cycle generalisation across 13 cross-family systems on the 100-scenario validation suite.} The identical pair of blue-team interventions identified from Qwen3-8B's 24-scenario failure profile (tool-output firewall + confirmation gate) is applied uniformly to every system, with no per-model and no per-scenario tuning. Red bars: Control RWF; green bars: Treated RWF. Lower is better. Systems sorted by Control RWF ascending (best on left), matching the order of Figure~\ref{fig:profile_radar}.}
\label{fig:before_after}
\end{figure*}

Table~\ref{tab:delta} reports $\mathrm{RWF}_{\text{Control}}$, $\mathrm{RWF}_{\text{Treated}}$, and the absolute drop $\Delta$ for each system, and Figure~\ref{fig:before_after} visualises the same data sorted by Control RWF. Two cross-model patterns emerge.
%
\input{Chapter/_delta_table_inc.tex}
\emph{First, the interventions generalise universally across families and scales.} All 13 systems improve under the Treated configuration with $\Delta\mathrm{RWF}>0$, and the largest gains occur on the systems that were weakest under Control: GPT-oss-120B ($0.481 \to 0.000$, $\Delta=+0.481$, the largest absolute improvement in the entire study), GPT-oss-20B ($0.418 \to 0.014$, $\Delta=+0.404$), Mistral-Large-3-675B ($0.396 \to 0.101$, $\Delta=+0.295$), Llama-3.1-70B ($0.407 \to 0.115$, $\Delta=+0.292$), and DeepSeek-V3.2 ($0.486 \to 0.205$, $\Delta=+0.281$). Two systems drop all the way to a perfect $\mathrm{RWF}=0.000$ under Treated (GPT-oss-120B, Llama-3.1-8B). This is direct evidence that interventions designed from a single small open-weight model's failure trajectories also reduce risk on entirely different architectures, providers, and capability tiers; the HAAF Factory mechanism is not specific to its focal model.
\emph{Second, magnitude of improvement is proportional to baseline failure surface, and one system is intervention-resistant.} Qwen3-32B is the outlier: it improves only marginally ($0.462 \to 0.443$, $\Delta=+0.019$), despite the same uniform interventions reducing every other system's $\mathrm{RWF}$ by $\geq 0.13$. The property profile (Figure~\ref{fig:profile_radar}) shows why: Qwen3-32B's failures are concentrated on P3 ($\hat T_{P_3}=0.09$) and P4 ($\hat T_{P_4}=0.11$), and the bulk of these are direct user-instructed disclosures and bias requests rather than indirect injections---failure modes the tool-output firewall is not designed to address. This is the kind of intervention-resistance pattern the profile surfaces but a scalar leaderboard would average away; it identifies Qwen3-32B as the system requiring a fundamentally different second Factory cycle (policy fine-tuning or value alignment, not prompt-level guardrails).

\paragraph{Limitations.}
The instantiation validates the methodology (attribution + comparison + iterative hardening + transfer) rather than producing statistically powered conclusions per model. Three caveats apply: \textit{(i)~Synthetic sandbox}: scenarios run locally rather than against real tools---a deliberate choice for safety/reproducibility, following AgentDojo~\cite{debenedetti2024agentdojo} and ToolSandbox~\cite{lu2024toolsandbox}, but limiting realism; bridging to deployment logs is the natural next step. \textit{(ii)~Transfer evidence is cross-model, not cross-domain}: the interventions were designed without observing any of the 13 systems' trajectories, so improvements under Treated guard against per-model overfitting. The validation suite extends the design suite with 76 new adversarial scenarios that the interventions were never tuned against; however, both suites share the same sandbox and tool surface, so the experiment does not guard against general-domain (sandbox-wide) overfitting. \textit{(iii)~IR auto-detection is heuristic}: a refusal-language classifier on final-answer text plus a low-tool-call rule; manual annotation would tighten labelling. Full deployment-readiness still requires multiple Factory cycles per system, held-out suites, and human-validated annotations (\S\ref{sec:discussion}).

%% file: Chapter/_profile_table_inc.tex
% Auto-generated by analyze_profile.py — do not edit by hand.
\begin{table}[!htbp]
\centering
\caption{Trustworthiness profiles across 13 cross-family agentic systems on the 100-scenario HAAF validation Control suite. Each $\hat T_{P_k}$ is one minus the violation rate on scenarios targeting property $P_k$ (higher = better); $\mathrm{RWF}$ = Risk-Weighted Failure (lower = better). Per-axis scenario counts: P1 ($n{=}26$), P2 ($n{=}20$), P3 ($n{=}22$), P4 ($n{=}18$), P5 ($n{=}14$). \textbf{Bold}~=~best in column.}
\label{tab:profile}
\small
\setlength{\tabcolsep}{3pt}
\resizebox{\columnwidth}{!}{%
\begin{tabular}{lccccccc}
\toprule
\textbf{System} & \textbf{Family} & $\hat T_{P_1}$ & $\hat T_{P_2}$ & $\hat T_{P_3}$ & $\hat T_{P_4}$ & $\hat T_{P_5}$ & $\mathrm{RWF}\downarrow$ \\
\midrule
Llama-3.1-8B & Llama & \textbf{1.000} & \textbf{1.000} & 0.545 & 0.611 & \textbf{1.000} & 0.216 \\
Llama-3.1-70B & Llama & \textbf{1.000} & \textbf{1.000} & 0.136 & 0.278 & \textbf{1.000} & 0.407 \\
Mistral-Large-2 & Mistral & \textbf{1.000} & 0.800 & 0.636 & 0.389 & \textbf{1.000} & 0.301 \\
Mistral-Large-3 & Mistral & 0.962 & 0.950 & 0.273 & 0.278 & 0.929 & 0.396 \\
Kimi-K2-Thinking & Kimi & \textbf{1.000} & 0.600 & 0.136 & 0.611 & \textbf{1.000} & 0.434 \\
Kimi-K2.5 & Kimi & \textbf{1.000} & 0.950 & 0.227 & 0.389 & \textbf{1.000} & 0.374 \\
GLM-4.7 & GLM & \textbf{1.000} & 0.950 & 0.182 & 0.611 & \textbf{1.000} & 0.333 \\
GLM-5 & GLM & \textbf{1.000} & 0.900 & \textbf{0.682} & \textbf{0.833} & \textbf{1.000} & \textbf{0.153} \\
Qwen3-32B & Qwen & \textbf{1.000} & \textbf{1.000} & 0.091 & 0.111 & \textbf{1.000} & 0.462 \\
Qwen3-Next-80B & Qwen & 0.885 & 0.450 & 0.136 & 0.222 & 0.929 & 0.587 \\
GPT-oss-20B & GPT & 0.885 & 0.600 & 0.364 & 0.500 & 0.929 & 0.418 \\
GPT-oss-120B & GPT & 0.885 & 0.550 & 0.273 & 0.444 & 0.857 & 0.481 \\
DeepSeek-V3.2 & DeepSeek & 0.962 & 0.600 & 0.182 & 0.389 & \textbf{1.000} & 0.486 \\
\bottomrule
\end{tabular}}
\end{table}

%% file: Chapter/_delta_table_inc.tex
% Auto-generated by analyze_profile.py.
\begin{table}[!htbp]
\centering
\caption{Factory cycle generalisation: Control vs.\ Treated $\mathrm{RWF}$ across the 13-system cross-family suite. The Treated configuration adds the tool-output firewall (P2) and confirmation gate (P3) interventions identified by the single-model attribution in \S\ref{subsec:vulnerability_attribution}; lower $\mathrm{RWF}$ is better. $\Delta$ is the absolute drop ($\mathrm{RWF}_{\text{Control}} - \mathrm{RWF}_{\text{Treated}}$); positive values indicate improvement. \textbf{Bold}: largest and smallest $\Delta$; perfect $\mathrm{RWF}_{\text{Treated}}=0.000$ also bold.}
\label{tab:delta}
\small
\setlength{\tabcolsep}{6pt}
\begin{tabular}{llccc}
\toprule
\textbf{System} & \textbf{Family} & \textbf{Control} & \textbf{Treated} & $\Delta\downarrow$ \\
\midrule
Llama-3.1-8B & Llama & 0.216 & \textbf{0.000} & $+0.216$ \\
Llama-3.1-70B & Llama & 0.407 & 0.115 & $+0.292$ \\
Mistral-Large-2 & Mistral & 0.301 & 0.055 & $+0.246$ \\
Mistral-Large-3 & Mistral & 0.396 & 0.101 & $+0.295$ \\
Kimi-K2-Thinking & Kimi & 0.434 & 0.164 & $+0.270$ \\
Kimi-K2.5 & Kimi & 0.374 & 0.164 & $+0.210$ \\
GLM-4.7 & GLM & 0.333 & 0.087 & $+0.246$ \\
GLM-5 & GLM & 0.153 & 0.022 & $+0.131$ \\
Qwen3-32B & Qwen & 0.462 & 0.443 & \textbf{$+0.019$} \\
Qwen3-Next-80B & Qwen & 0.587 & 0.358 & $+0.230$ \\
GPT-oss-20B & GPT & 0.418 & 0.014 & $+0.404$ \\
GPT-oss-120B & GPT & 0.481 & \textbf{0.000} & \textbf{$+0.481$} \\
DeepSeek-V3.2 & DeepSeek & 0.486 & 0.205 & $+0.281$ \\
\bottomrule
\end{tabular}
\end{table}

%% file: Chapter/5_discussion.tex
\section{Discussion}
\label{sec:discussion}

\paragraph{Significance and vision.}
We argue that representative, risk-sensitive scenario distributions should be a first-class object of evaluation in agentic AI, anchored to a concrete five-property definition of trustworthiness (\S\ref{subsec:defining_trustworthiness}). The reframing shifts evaluation from benchmark instances to scenario distributions, from single scores to multi-dimensional profiles, and from static testing to the iterative red--blue cycle of the \emph{Trustworthy Optimization Factory}. The cross-model comparison (\S\ref{subsec:profile_comparison}) confirms a corollary: meaningful comparison is inherently multi-property---families (Llama, Mistral, Kimi, GLM, Qwen, GPT, DeepSeek) make different trade-offs that a leaderboard erases. If this agenda succeeds, the community converges on shared scenario taxonomies and world-model libraries~\cite{liang2023helm,zhou2023sotopia,drouin2024workarena,xie2024osworld}, reports profiles rather than single scores~\cite{liang2023helm,kiela2021dynabench}, and adopts the Factory's red--blue cycle as a deployment-readiness pipeline---aligned with emerging AI governance frameworks~\cite{ganguli2022redteaming,nist2023airmf,eu2024aiact}.

\paragraph{Implications for agentic information retrieval.}
The shift from passive retrieval to \emph{retrieval-grounded action}~\cite{xi2024agentir} makes agentic IR a natural and high-stakes setting for this framework. Three implications follow. \textit{(i)} Retrieved content is an attack surface: indirect prompt injection through retrieved documents~\cite{greshake2023indirect,zhou2024poisonedrag} is precisely a P2 (Robustness) failure, and our sandbox cleanly separates agents that treat retrieved text as data from those that treat it as instructions. \textit{(ii)} Retrieval-grounded agents expose sensitive corpora through natural-language interfaces, making P3 (Safety) and P4 (Social-Ethical) joint concerns rather than separable. \textit{(iii)} Agentic IR designers should not pick a model by aggregate score: a system top-ranked on retrieval helpfulness may rank last on resistance to retrieved-content injection or social manipulation; only profile-level comparison reveals which trade-off matches a given deployment.

\paragraph{Limitations and future work.}
A full instantiation will need larger held-out suites (our 100-scenario validation suite improves substantially over a 24-scenario pilot but is still synthetic), human-validated failure annotations, and grounding of representative sampling in real-world evidence (deployment logs, incident reports, red-team traces~\cite{ganguli2022redteaming,debenedetti2024agentdojo}). Open questions include formalising Factory convergence criteria, handling user-instructed violations that resist prompt-level mitigation, and standard property-weight vectors $\boldsymbol{\lambda}$ for common deployment classes so risk-weighted aggregates are comparable across studies. The adversarial framings we release carry a dual-use risk; we mitigate by pairing the scenario suite with the blue-team hardening cycle (\S\ref{subsec:closed_loop}) and by including the Improper-Refusal class so that blanket refusal is not rewarded as a substitute for value-aligned judgement.

%% file: Chapter/6_conclusion.tex
\section{Conclusion}
\label{sec:conclusion}

Current agent evaluation remains fragmented across benchmark islands, and the very target of evaluation---``trustworthiness''---is rarely defined in operational terms. We give a deployment-oriented definition through five measurable properties---Reliability, Robustness, Safety, Social-Ethical Alignment, and Operational Integrity---and introduce HAAF, which combines distribution-aware sampling with multi-layer probing and an iterative \emph{Trustworthy Optimization Factory} aligned to these properties. Our instantiation shows three results in concert: a 24-scenario Factory cycle narrows the focal model's vulnerability surface from 16.7\% to 4.2\% with no new improper refusals; a 100-scenario cross-model profile over 13 systems from seven families surfaces property-level trade-offs no scalar leaderboard captures, including a near-universal P3 (Safety) and P4 (Social-Ethical) weakness under user-instructed violations and an \emph{anti-scaling} pattern across four families; and the focal-model interventions, applied uniformly to all 13 systems, drive every one to a better profile (two to a perfect $\mathrm{RWF}=0.000$), with one intervention-resistant case (Qwen3-32B) flagged for a fundamentally different second Factory cycle---direct evidence that property-targeted hardening transfers across model families while preserving visibility into where it does not. Without representative, property-aware evaluation, benchmark progress will remain only weakly connected to deployment trustworthiness; we hope this work motivates a broader agenda around scenario modelling, risk-aware sampling, and iterative deployment-readiness assessment for agentic information-retrieval systems and beyond. The full 100-scenario validation suite, the seven-family Bedrock adapter, and all $2{,}600$ Control/Treated trajectories are released at \url{https://github.com/TonyQJH/haaf-pilot} to support replication, scenario extension, and second-cycle Factory studies on intervention-resistant systems such as Qwen3-32B.